\documentclass[runningheads]{llncs}
\usepackage{graphicx}
\usepackage{comment}
\usepackage{amsmath,amssymb} 
\usepackage{color}
\usepackage{multirow}
\usepackage[colorlinks,linkcolor=red]{hyperref}

\usepackage{eso-pic}
\usepackage{amssymb}

\begin{document}
\pagestyle{headings}
\mainmatter
\def\ECCVSubNumber{981}  

\title{Discriminability Distillation \\
in Group Representation Learning} 

\titlerunning{Discriminability Distillation
in Group Representation Learning}
%
\author{Manyuan Zhang\inst{1, 2} \and
Guanglu Song\inst{1} \and
Hang Zhou\inst{2} \and Yu Liu\inst{1, 2}\thanks{Corresponding author.}}
\authorrunning{Manyuan Zhang, Guanglu Song, Hang Zhou, Yu Liu}
%
\institute{SenseTime X-Lab\\
\and
CUHK - SenseTime Joint Lab, The Chinese University of Hong Kong \\
\email{\{zhangmanyuan,songguanglu\}@sensetime.com} \\
\email{zhouhang@link.cuhk.edu.hk},
\email{liuyuisanai@gmail.com }
}
\maketitle

\begin{abstract}
Learning group representation is a commonly concerned issue in tasks where the basic unit is a group, set, or sequence.
Previously, the research community tries to tackle it by aggregating the elements in a group based on an \textit{indicator} either defined by humans such as the \textit{quality} and \textit{saliency}, or generated by a black box such as the attention score. This article provides a more essential and explicable view. 
We claim the most significant indicator to show whether the group representation can be benefited from one of its element is not the quality or an inexplicable score, but the \textit{discriminability w.r.t.} the model. 
We explicitly design the \textit{discriminability} using embedded class centroids on a proxy set. We show the discriminability knowledge has good properties that can be distilled by a light-weight distillation network and can be generalized on the unseen target set.
The whole procedure is denoted as \textit{discriminability distillation learning} (DDL).
The proposed DDL can be flexibly plugged into many group-based recognition tasks without influencing the original training procedures. Comprehensive experiments on various tasks have proven the effectiveness of DDL for both accuracy and efficiency. Moreover, it pushes forward the state-of-the-art results on these tasks by an impressive margin. Code and models are available at \url{https://github.com/manyuan97/DDL/}.


\end{abstract}

\section{Introduction}
With the rapid development of deep learning and easy access to large-scale group data, recognition tasks using group information have drawn great attention in the computer vision community. The rich information provided by different elements can complement each other to boost the performance of tasks such as face recognition, person re-identification, and action recognition~\cite{wang2017untrimmednets,zhong2018ghostvlad,girdhar2017actionvlad,simonyan2014two,yang2017neural,liu2019permutation,rao2017attention}. For example, recognizing a person through a sequence of frames is expected to be more accurate than watching only one image.

While traditional practice for group-based recognition is to either aggregate the whole set by average pooling~\cite{li2014eigen,taigman2014deepface}, max pooling~\cite{chowdhury2016one}, or just randomly sampling ~\cite{wang2016temporal}, the fact that certain elements contribute negatively in recognition tasks has been ignored. Thus, the key problem for group-based recognition is how to define an efficient  indicator to select representatives from sets.



To tackle such cases, previous methods aim at defining the “quality” or “saliency” for each element in a group~\cite{liu2017quality,yang2017neural,rao2017attention,nikitin2017neural}. The weight for each element can be automatically learned by self-attention. For example, Liu et al.~\cite{liu2017quality} propose the Quality Aware Network (QAN) to learn a quality score for each image inside an image set during network training. Other researchers adopt the same idea and extend it to specific tasks such as video-based person re-identification~\cite{Li_2018_CVPR,wu2018and} and action recognition~\cite{wang2018non} by learning spatial-temporal attentions. However, the whole quality/attention learning procedures are either manually designed or learned through a black box, which lacks explainability. Moreover, since previous attention and quality mechanism are mostly based on element feature, the features for all group elements need to be extracted, which is highly computational consuming.

In this work, we explore deeper into the underlying mechanism for defining effective elements. Assuming that a base network $\mathcal{M}$ has already been trained for element-based recognition using class labels, we define the ``discriminability'' of one element by how difficult it is for the network $\mathcal{M}$ to discriminate its class. How to measure the difficulty and the learning preference of the network $\mathcal{M}$ of elements remains an interesting problem. By considering the relationship between intra- and inter-class distance, we identify a successful discriminability indicator by \emph{measuring one embedding’s distance with all class centroids and compute the ratio of between positive and hardest-negative}. The \emph{positive} is its distance from its class's corresponding centroid and the \emph{hardest-negative} is its closest counterpart.


%


As the acquiring procedure of the discriminability indicator is highly flexible without either human supervision or network re-training, it can be adapted to any existing base. Though defined through trained bases, we find that the discriminability indicator can be easily distilled by training an additional light-weight network (Discriminability Distillation Network, DDNet). The DDNet takes the raw images as input and regresses the regularized discriminability indicators. We uniformly call the whole procedure \emph{discriminability distillation learning} (DDL).

During inference, all elements are firstly sent into the light-weight DDNet to estimate their discriminability. Then element features will be weighted and aggregated according to their discriminability scores. In addition, in order to achieve the trade-off between accuracy and efficiency, we can filter elements by extracting and aggregating elements of high discriminability only. Since the base model tends to be heavy, the filtering process can save much computational cost. We evaluate the effectiveness of our proposed DDL on several classical yet challenging tasks including set-to-set face recognition, video-based person re-identification, and action recognition. Comprehensive experiments show the advantage of our method on both recognition accuracy and computational efficiency. State-of-the-art results can be achieved without modifying the base network.


We highlight our contributions as follows: (1) We define the \emph{discriminability} of one element within a group from a more essential and explicable view, and propose an efficient indicator.  Moreover, we demonstrate that the structure of discriminability distribution can be easily distilled by a light-weight network. (2) With a well-designed element discriminability learning and feature aggregating process, both efficiency and excellent performance can be achieved. We verify the good generalization ability of our discriminability distillation learning  in many group-based recognition tasks, including set-to-set face recognition, video-based person re-identification, and action recognition through extensive studies. 


\section{Related work}

\setcounter{secnumdepth}{2}

Group representation learning which aims at formulating a unified representation has been proved efficient on various tasks~\cite{zhong2018ghostvlad,liu2017quality,gao2018revisiting,wang2016temporal,zhou2019talking}. In this paper, we care for three group representation learning tasks including set-to-set face recognition, video-based person re-identification, and action recognition. In this section, we will briefly review those related topics.


\noindent{\textbf{Set-to-Set Face Recognition.}} Set-to-set face recognition aims at performing face recognition~\cite{wolf2011face,kalka2018ijb,beveridge2013challenge,klare2015pushing,deng2019arcface,Zhou_2020_CVPR} using a set of images of a same person.
To tackle set-to-set face recognition, traditional methods directly estimate the feature similarity among sets of feature vectors~\cite{arandjelovic2005face,harandi2011graph,cevikalp2010face}. Other works seek to aggregate element features by simply applying max-pooling~\cite{chowdhury2016one} or average pooling~\cite{li2014eigen,taigman2014deepface} among set features to form a compact representation. However, since most set images are under unconstrained scenes, huge variations such as blur and occlusions will degrade the set feature discrimination. How to design a proper aggregation method for set face representation has been the key. 

Recently, a few methods explore the manually defined operator or attention mechanism to form group representation. GhostVLAD~\cite{zhong2018ghostvlad} improves traditional VLAD. While Rao \textit{et al.}~\cite{rao2017learning} combine LSTM and reinforcement learning to discard low-quality element features. Liu \textit{et al.}~\cite{liu2017quality} and Yang \textit{et al.}~\cite{yang2017neural} introduce an attention mechanism to assign quality scores for different elements and aggregate feature vectors by quality weighted sum. To predict the quality score, an online attention network module is added and co-optimized by the target set-to-set recognition task.  However, the definition of generated ``quality'' scores remains unclear and they are learned through a black box, which lacks explainability.

\noindent{\textbf{Video-Based Person Re-Identification.}} It is also beneficial to perform person re-identification~\cite{yan2016person,mclaughlin2016recurrent,ge2018fd,ge2020mutual,ge2020self,Li_2018_CVPR,gao2018revisiting} from videos. There are typically three components for video-based person re-identification: an image-level feature extractor, a temporal aggregating module, and the loss function~\cite{gao2018revisiting}. Previous works mainly focus on optimizing the temporal aggregating module for video-based person re-identification. They can be divided into three categories, RNN-based~\cite{mclaughlin2016recurrent,yan2016person}, attention-based~\cite{liu2017quality,zhou2017see} and 3D-Conv based~\cite{gao2018revisiting}. Yang \textit{et al.}~\cite{yan2016person} model an RNN to encode element features and use the final hidden layer as the group feature representation. Liu \textit{et al.}~\cite{liu2017quality} use attention module to assign each element an quality score. While Gao \textit{et al.}~\cite{gao2018revisiting} directly utilize 3D Conv to encode the spatial-temporal feature for elements and propose a benchmark to compare different temporal aggregating module fairly.

\noindent{\textbf{Action Recognition.}} Action representation learning is another typical case of group-based representation learning. Real-world videos contain variable frames, so it is not practical to put the whole video to a memory limited GPU. The most usual approach for video understanding is to sample frames or clips and design late fusion strategies to form the video-level prediction.

Frame-based methods~\cite{yue2015beyond,gan2018geometry,simonyan2014two,girdhar2017actionvlad} firstly extract  frame features and aggregate them. Simonyan~\textit{et al.}~\cite{simonyan2014two} propose the two-stream network to simultaneously capture the appearance and motion information. Wang~\textit{et al.}~\cite{wang2017untrimmednets} add attention module and learn to discard unrelated frames. Frame-based methods are computationally efficient, but only aggregating high-level frame features tends to limit the model's ability to handle complex motion.

Clip-based methods~\cite{tran2015learning,tran2018closer,feichtenhofer2018slowfast,korbar2019scsampler}  use 3D convolutional neural network to jointly capture spatial-temporal features. However, clip-based methods highly rely on the dense sample strategy, which introduces huge computational costs and makes it impractical to real-world applications. In this article, we show that by combining our DDL, the clip-based methods can achieve both excellent performance and computational efficiency.

\section{Discriminability Distillation Learning}

In this section, we first formulate the problem of group representation learning in section \ref{sub:3_1} and then define the discriminability in section \ref{sub:3_2}. Next, we introduce the whole discriminability distillation learning (DDL) procedure in section \ref{sub:3_3}. In sections \ref{sub:3_4} and \ref{sub:3_5}, we discuss the aggregation method and the advantage of our DDL, respectively.

\subsection{Formulation of Group Representation Learning}
\label{sub:3_1}


Compared to using a single element, performing recognition with group representation can further explore the complementary information among group elements and benefit from them. 
For example, recognizing a person from a group of his photos instead of one image is sure to facilitate the result. 

The most popular way to handle group-based recognition tasks is to formulate a unified representation for a whole group of elements~\cite{liu2017quality,zhong2018ghostvlad,wang2016temporal,gao2018revisiting}. 
%
Suppose a base network $\mathcal{M}$ is trained for the element-based recognition task. 
Define $f_{i}\in \mathbb{R}^{d}$ as the embedded feature of element $I_{i}$ in group ${\bf{I}}_{S}$ from $\mathcal{M}$, the unified feature representation of the whole group is
\begin{equation}
f_{{\bf{I}}_{S}}=\mathcal{G}(f_{1},f_{2},\cdots,f_{i}),
\label{eq:FIS}
\end{equation}
where $\mathcal{G}$ indicates the feature aggregation module.
While previous research has revealed that conducting $\mathcal{G}$ with quality~\cite{liu2017quality} has priority over simple aggregation, this kind of method is not explainable and computation-consuming. In this article,
we propose discriminability distillation learning (DDL) to generate the \emph{discriminability} of feature representation.

\begin{figure}[t]
\begin{center}
\includegraphics[width=0.7\linewidth]{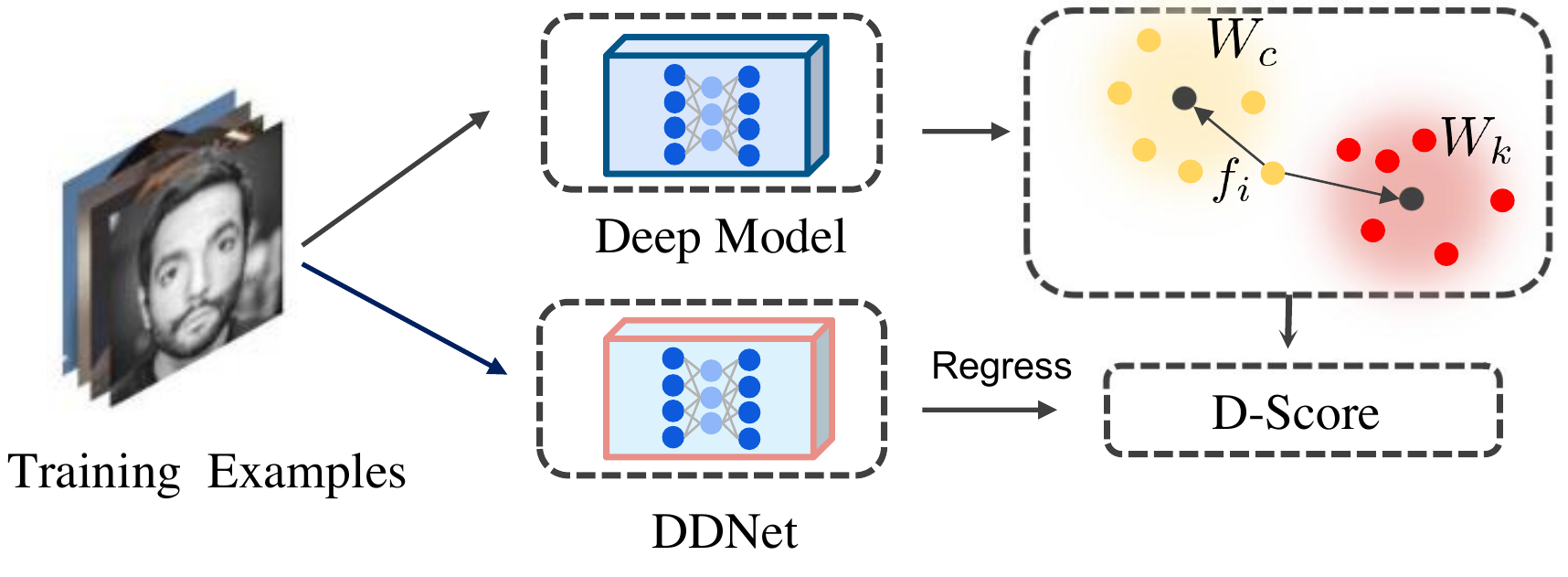}
\end{center}
   \caption{The pipeline of group representation learning with  DDL. Given a base feature extracting model, we first compute the discriminability  for each training element and then train a light-weight discriminability distillation network  (DDNet) to regress it. The discriminability  is formulated from the view of intra and inter-class distance with class centroids for element}
\label{fig:pic1}
\end{figure}

\subsection{ Formulation of Discriminability }
\label{sub:3_2}

Towards learning efficient and accurate $\mathcal{G}$, we propose to define the \emph{discriminability} of elements to replace the traditional quality or attention mechanism. 

After training the base model $\mathcal{M}$ on the classification task, features of the training elements from the same class are projected to hyperspace tightly in order to form an implicit decision boundary and minimizing target loss~\cite{liu2018transductive}. This statement exists when $\mathcal{M}$ is supervised by all kinds of loss functions (softmax-cross entropy~\cite{sun2014deep}, triplet~\cite{schroff2015facenet} or margin-based~\cite{deng2019arcface,duan2019uniformface} losses). Our key observation is that the features embedded close to their corresponding class centroids are normally the representative examples, while features far away  or closer to other centroids are usually the confusing ones.

Based on our motivation, we jointly consider the feature space distribution and explicitly distill the \emph{discriminability} by encoding the intra-class distance and inter-class distance with class centroids.
Let $\mathcal{X}$ denotes the training set with $K$ classes and $C_{m}\in \mathbb{R}^{d}, m \in[1, K]$ is the class centroid of class $m$, which is the average of features. For feature $f_{i}, i\in [1,s]$ where $s$ denotes the size of $\mathcal{X}$. Assume the positive class for $f_{i}$ is $p$, while the negatives are $n\in [1, K], n\neq p $. The intra-class distance and inter-class distance  for $f_{i}$ are formulated as:
 \begin{equation}
 \begin{split}
    {dist}_{ip} &= \frac{f_i \cdot C_p}{\| f_i\|_2\ \| C_p\|_2}, \\
    {dist}_{in} &= \frac{f_i \cdot C_n}{\| f_i\|_2\ \| C_n\|_2},\ n\in [1, K], n\neq p.
\end{split}
\label{eq:C}
\end{equation}
Here we use the cosine distance as feature distance metric. Other metrics like Euclidean distance are also applicable. 
Then the \emph{discriminability} $\mathcal{D}_i$ of $f_i$ can be defined as:
 \begin{equation}
\mathcal{D}_{i}=\frac{dist_{ip}}{\max \left\{dist_{in} \;| \ n \in[1, K], n \neq p\right\}}.
\label{eq:Qi}
\end{equation}
It is the ratio between the feature's distance from the centroid of its own class and the distance from  the hardest-negative class. Considering the variant number of elements in different groups, we further normalize the \emph{discriminability} by:
 \begin{equation}
\mathcal{D}_{i}=\tau\left(\frac{\mathcal{D}_{i}-\mu(\{\mathcal{D}_j \;|\ j\in[1,s]\})}{\sigma(\{\mathcal{D}_j \;|\ j\in[1,s]\})}\right)
\label{eq:Di}
\end{equation}
 where $\tau(\cdot)$, $\mu(\cdot)$ $\text { and }$ $\sigma(\cdot)$ denote the sigmoid function, the mean value and the standard deviation value of $\{\mathcal{D}_j \;|\ j\in[1,s]\}$, respectively. We denote  the  normalized  $\mathcal{D}_i$ as  discriminability score (D-score).
 
Cooperated with the feature space distribution, the  discriminability $\mathcal{D}_i$ is more interpretable and reasonable. It can discriminate features better by explicitly encoding the intra- and inter-class distances with class centroids.

\subsection{Discriminability Distillation Learning}
\label{sub:3_3}

From section \ref{sub:3_2}, given a  base model $\mathcal{M}$ and its training dataset, the  $\mathcal{D}_i$ of $f_i$ can be naturally computed by Eq (\ref{eq:C})-(\ref{eq:Di}).
However, the score is unavailable to test set $\mathcal{T}$. In order to estimate unseen element's discriminability, we formulate the discriminability distillation learning (DDL) procedure for group representation.

Our idea is to \emph{distill} the discriminability explicitly using a light-weight auxiliary network from the training samples.
It is called the Discriminability Distillation Network (DDNet). Denote the  DDNet as $\mathcal{N}$, the approximated $\hat{\mathcal{D}_i}$  for $\mathcal{D}_i$  can be  given by: 
 \begin{equation}
\hat{\mathcal{D}}_{i}= \mathcal{N}(I_i ; \boldsymbol{\theta}),
\label{eq:hatDi}
\end{equation}
where $\boldsymbol{\theta}$ denotes the parameters of    $\mathcal{N}$.
To train $\mathcal{N}$, we apply mean squared error between $\hat{\mathcal{D}_i}$ and target $\mathcal{D}_i$ as
 \begin{equation}
L = \frac{1}{2N} \sum^N_i (\hat{\mathcal{D}_i}-\mathcal{D}_i)^2,
\end{equation}
where $N$ is the batch size. The training  is conducted with the same training set for the base model and there is no need to modify the base model $\mathcal{M}$.

\subsection{Feature Aggregation $\mathcal{G}$}
\label{sub:3_4}

During inference, we can generate  $\hat{\mathcal{D}_i}$ via Eq (\ref{eq:hatDi}) for each element $I_i$ in the given element set $I_S$. Then we can filter out some elements with low  discriminability in order to accelerate the feature extracting process of $\mathcal{M}$. Given the pre-defined  threshold $t$ and base model $\mathcal{M}$, the group element feature extracting process is
\begin{equation}
f_{i}= \mathcal{M}(I_i), \hat{\mathcal{D}_i} > t,
\end{equation}
and $\mathcal{G}$ in Eq (\ref{eq:FIS}) can be formulated as:
\begin{equation}
F_{{\bf{I}}_{S}}= \mathcal{G}(f_{1},f_{2},\cdots,f_{n}) =\sum_{i}^{n} \frac{\hat{\mathcal{R}_i} f_{i}}{\hat{\mathcal{R}_i}},
\end{equation}
where $n$ is the  number of $I_S$ whose discriminability is higher than threshold $t$, and $\hat{\mathcal{R}_i}$ is the re-scaled D-score  via 
\begin{figure}[t]
\begin{center}
\includegraphics[width=0.8\linewidth]{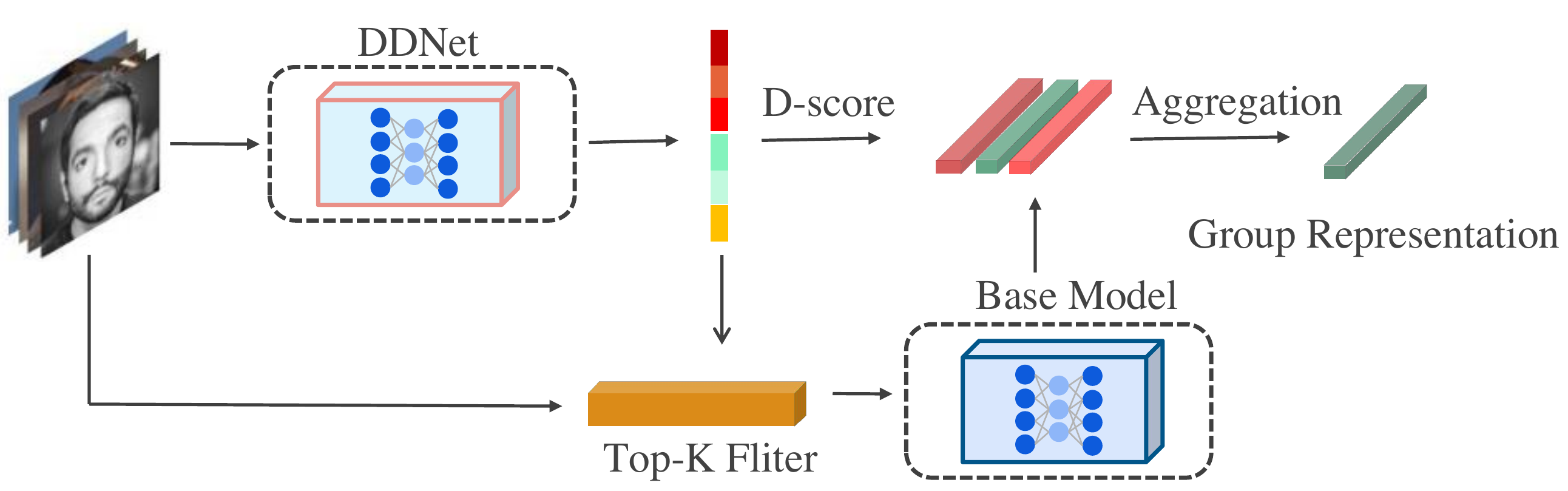}
\end{center}
   \caption{The pipeline of the test stage with DDL. For a group of elements, we first predict D-score by the trained-well DDNet for each element. Then we will filter elements by their D-scores and only extract feature for those elements with high D-scores by the base model. Finally extracted features will be weighted sum to form the group representation
}
\label{fig:pic2}
\end{figure}

\begin{equation}
\hat{\mathcal{R}_i}=K \hat{\mathcal{D}_i}+B.
\label{eq:hatRi}
\end{equation}

In Eq (\ref{eq:hatRi}), we scale the D-score of  element set $I_S$ between 0 and 1 to ensure the same range for element sets with different lengths.
$K$ and $B$ are formulated as

\begin{equation}
K=\frac{1}{\max \{\hat{\mathcal{D}_i}\;|\ i \in [1,n]\} - \min\{\hat{\mathcal{D}_i}\;|\ i \in [1,n]  \} },
\end{equation}
\begin{equation}
B=1-K\max \{\hat{\mathcal{D}_i}\;|\ i \in [1,n]\}.
\end{equation}

\subsection{Advantage of Discriminability Distillation Learning}
\label{sub:3_5}

Different from the subjective quality judgment of an image or the attention mechanism, we explicitly assign \emph{discriminability} for each element via the feature space distribution. 
By jointly considering the inter- and intra-class distances with class centroids, DDL can effectively approximate how discriminative a feature is. 
By aggregating more information with features with high \emph{discriminability}, more discriminative group representation can be formed, leading to a significant performance boost for group-based recognition tasks.
In addition, the well-design discriminability distillation learning process needn't modify the base model, making it easy to be plugged into many popular recognition frameworks.
Furthermore, We can change the threshold for the discriminability filtering process according to different application scenarios to achieve a trade-off between accuracy and computational cost.

\section{Experiments}

We evaluate our DDL on three popular group-based recognition tasks: set-to-set face recognition, video-based person
re-identification, and action recognition. An ablation study will be conducted along with the set-to-set face recognition experiments.

\subsection{Set-to-Set Face Recognition}

In this section, we evaluate DDL for set-to-set face recognition on four datasets including two video datasets:  YouTube Face  (YTF)~\cite{wolf2011face}, iQIYI-VID-FACE~\cite{iQIYI-VID-FACE}; and two template-based datasets: IARPA Janus Benchmark A (IJB-A) ~\cite{klare2015pushing} and  IARPA Janus Benchmark C (IJB-C).

\setcounter{secnumdepth}{3}
\subsubsection{Implementation Details.}

For data pre-processing, RetinaFace~\cite{deng2019retinaface} is used to detect faces and their corresponding landmarks for all datasets. Images are aligned to $112\times 112$ by similarity transformation with facial landmarks.

We train our base model and DDNet on the MS-Celeb-1M dataset~\cite{guo2016ms} cleaned by~\cite{deng2019arcface}. The base model we select is modified ResNet-101~\cite{he2016deep} released by~\cite{deng2019arcface}. As for the DDNet, we use a light-weight channel reduced ResNet-18 network, whose channels for 4 stages are \{8, 16, 32, 48\}, respectively. It only introduces 81.9 Mflops, which is super-efficient. 

The loss function for the base model training is  ArcFace~\cite{deng2019arcface} and the total training step is 180k with initial learning rate 0.1 on 8 NVIDIA  Tesla V100 GPUs. The training process for our DDNet is similar to the base model. The default discriminability threshold we select is 0.15, empirically.

\subsubsection{Evaluation on YouTube Face.}

The YouTube Face~\cite{wolf2011face} dataset includes 3425 videos of 1595 identities with an average of 2.15 videos per identity. The videos vary from 48 frames to 6,070 frames.  We report the 1:1 face verification accuracy of the given 5,000 video pairs in our experiments.

As shown in Table \ref{tab:ytf}, our DDL achieves state-of-the-art performance on the YouTube Face benchmark~\cite{wolf2011face}. It outperforms~\cite{deng2019arcface} by 0.16\% and other set-to-set face recognition methods by impressive margins.  For comparison with different aggregation strategies like average pooling, DDL can boost performance by 0.21\%, which indicates DDL has learned a meaningful pattern for discriminability. As a post-training module, DDL can cooperate with any 
existing base. Note that if we only select the top-1 discriminability frame, DDL can also achieve 97.08\%, which achieves above 130x acceleration.  The computation complexity for the base model is 11 Gflops (ResNet-101) while our DDNet only introduces 81.9 Mflops. By filtering most frames, great computational cost is saved.

\subsubsection{Evaluation on IQIYI-VID-FACE.}

Since the results on YouTube Face benchmark tend to be saturated, we test our DDL on the challenging video face verification benchmark IQIYI-VID-FACE~\cite{deng2019lightweight}, 
The IQIYI-VID-FACE dataset aims to identify the person in entertainment videos by face images. It is the largest video face recognition test benchmark so far, containing 643,816 video clips of 10,034 identities. The test protocol is 1:1 verification, and the True Accept Rate (TAR) under False Accept Rate (FAR) at 1e-4 is reported.

As shown in Table \ref{tab:iqiyi}, compared with the average pooling, DDL improves performance by 3.21\%. Even only aggregating the top-1 discriminability score frame can still achieve an equal performance of average aggregation for all frames. It shows that our DDL has selected the most discriminative element of the set. By combining stronger base model PolyNet~\cite{zhang2017polynet}, our DDL achieves state-of-the-art performance on the IQIYI-VID-FACE challenge.

\begin{table}[t]
\caption{Video face verification performance on YouTube Face dataset, compared with state-of-the-art methods and baseline methods}
\centering

\setlength{\tabcolsep}{15pt}
\resizebox{1\textwidth}{!}{%
\renewcommand\arraystretch{1.1}
\begin{tabular}{cccc}\\ \hline
Method & Accuracy(\%) & Method & Accuracy(\%) \\ \hline
Li \textit{et al.}~\cite{li2014eigen} & 84.8 &DeepFace~\cite{taigman2014deepface} & 91.4 \\
FaceNet~\cite{schroff2015facenet} & 95.52 &NAN~\cite{yang2017neural} & 95.72 \\
DeepID2~\cite{sun2015deeply} & 93.20 & QAN~\cite{liu2017quality} & 96.17 \\
C-FAN~\cite{gong2019video} & 96.50 &Rao \textit{et al.}~\cite{rao2017attention} & 96.52 \\

Liu \textit{et al.}~\cite{liu2019feature} & 96.21 & Rao \textit{et al.}~\cite{rao2017learning} & 94.28 \\
CosFace~\cite{wang2018cosface} & 97.65 &ArcFace~\cite{deng2019arcface}& 98.02 \\ \hline
\textit{Average} & 97.97 & \textit{Top 1}  & 97.08 \\
 &  & DDL & \textbf{98.18} \\ \hline
\end{tabular}%
}
\label{tab:ytf}
\end{table}

\begin{table}[t]

\caption{Comparison with different participants and aggregation strategy on the IQIYI-VID-FACE challenge. By combining with PolyNet, DDL achieves state-of-the-art performance
}
\centering\centering
\setlength{\tabcolsep}{8pt}
\resizebox{1\textwidth}{!}{%

\renewcommand\arraystretch{1.1}
\begin{tabular}{cccc} \\ \hline
Method & TPR@FPR=1e-4(\%) & Method & TPR@FPR=1e-4(\%) \\ \hline
MSRA & 71.59 & Alibaba-VAG & 71.10 \\
Insightface & 67.00 & DDL (PolyNet)  & \textbf{72.98  } \\ \hline
\textit{Average} & 65.84 & \textit{Top 1} & 65.22 \\ DDL w/o re-scale
 & 67.38  & DDL  & \textbf{69.05} \\  \hline
\end{tabular}%
}
\label{tab:iqiyi}
\end{table}

\subsubsection{Evaluation on IJB-A and IJB-C.}

The IARPA Janus Benchmark A (IJB-A)~\cite{klare2015pushing} and IARPA Janus Benchmark C (IJB-C)  are unconstrained face recognition benchmarks. They are template-based test benchmarks where both still images and video frames are included in templates. IJB-A containing  25, 813 faces images of 500 identities while IJB-C has 140, 740 faces images of 3, 531 subjects. Since the images in IJB-C dataset have large variations, it is regarded as a challenging set-to-set face recognition benchmark.

Tables \ref{tab:ijba} and \ref{tab:ijbc} show the results on the IJB-A and IJB-C benchmark for different methods. From the two tables, we can see that our DDL improves verification performance by a convincing margin with average pooling for both two benchmarks, especially under severe FAR at 1e-5  by 0.73\% on IJB-A and FAR at 1e-6 by 5.7\% on IJB-C. Compared with IJB-A, IJB-C has more images and covers more variations among images, such as pose, blur, resolution, and conditions. So the performance gain with DDL is larger.

Compared with the state-of-the-art methods, our DDL improves IJB-A by 0.55\% when FAR =1e-3 and IJB-C by 6.14\% when FAR = 1e-6. These results indicate the effectiveness and robustness of our DDL. What's more, unlike many previous methods that need fine-tune with the base model on set-to-set recognition training datasets~\cite{liu2017quality,zhong2018ghostvlad}, the only supervision for DDL training is the  discriminability generated with the base model on the same training set, which is highly flexible.

\begin{table}[t]

\caption{Performance comparisons on IJB-A verification benchmark. The True Accept Rates (TAR) vs. False Accept Rate (FAR) are reported}
\centering
\setlength{\tabcolsep}{15pt}
\resizebox{1\textwidth}{!}{%
\renewcommand\arraystretch{1.1}
\begin{tabular}{cccc} \\ \hline
\multirow{2}{*}{Method} & \multicolumn{3}{c}{IJB-A (TAR@FAR)} \\ \cline{2-4} 
 & FAR=1e-3(\%) & FAR=1e-2(\%) & FAR=1e-1(\%) \\ \cline{1-4} 
Template Adaptation~\cite{crosswhite2018template} & 83.6 $\pm$ 2.7 & 93.9 $\pm$ 1.3 & 97.9 $\pm$ 0.4 \\
TPE~\cite{sankaranarayanan2016triplet}& 81.30 $\pm$ 2.0 & 91.0 $\pm$ 1.0 & 96.4 $\pm$ 0.5 \\
Multicolumn~\cite{xie2018multicolumn}& 92.0 $\pm$ 1.3 & 96.2 $\pm$ 0.5 & 98.9 $\pm$ 0.2 \\
QAN~\cite{liu2017quality} & 89.31 $\pm$ 3.92 & 94.2 $\pm$ 1.53 & 98.02 $\pm$ 0.55 \\
VGGFace2~\cite{cao2018vggface2} & 92.1 $\pm$ 1.4 & 96.8 $\pm$ 0.6 & 99 $\pm$ 0.2 \\
NAN~\cite{yang2017neural} & 88.1 $\pm$ 1.1 & 94.1 $\pm$ 0.8 & 97.8 $\pm$ 0.3 \\
GhostVLAD~\cite{zhong2018ghostvlad} & 93.5 $\pm$ 1.5 & 97.2 $\pm$ 0.3 & 99.0 $\pm$ 0.2 \\ 

Liu \textit{et al.}~\cite{liu2019feature} & 93.61 $\pm$ 1.51 & 97.28 $\pm$ 0.28 & 98.94 $\pm$ 0.31 \\
ArcFace~\cite{deng2019arcface} & 97.89 $\pm$ 1.5 & 98.51 $\pm$ 0.3 & 99.05 $\pm$ 0.2 \\ \hline
\textit{Average}  &  97.71 $\pm$ 0.6  & 98.43$\pm$ 0.4 & 99.01$\pm$ 0.2\\
DDL & \textbf{98.44 $\pm$ 0.3} & \textbf{98.79 $\pm$ 0.2} & \textbf{99.13 $\pm$ 0.1} \\ \hline

\end{tabular}%
}
\label{tab:ijba}
\end{table}

\begin{figure}[t]
\begin{center}
\includegraphics[width=\linewidth]{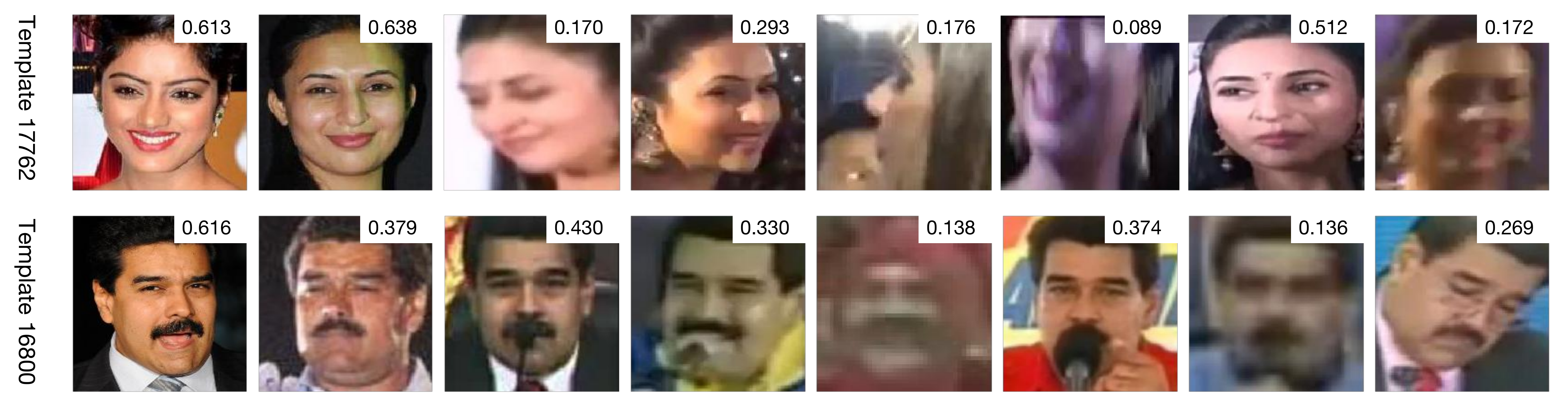}
\end{center}
  \caption{The visualization results of   discriminability   for images of  Template ID 17762 and 16800 from IJB-C dataset
}
\label{fig:vis_ijbc}
\end{figure}

\begin{table}[t]

\caption{Performance comparisons on IJB-C verification benchmark. The True Accept Rates (TAR) vs. False Accept Rate (FAR) are reported}
\centering
\setlength{\tabcolsep}{12pt}
\resizebox{.9\textwidth}{!}{%
\renewcommand\arraystretch{1.1}
\begin{tabular}{cccccc} \\ \hline
\multirow{2}{*}{Method} & \multicolumn{5}{c}{IJB-C (TAR@FAR)} \\ \cline{2-6} 
 & 1e-6(\%) & 1e-5(\%) & 1e-4(\%) & 1e-3(\%) & 1e-2(\%) \\ \hline
Yin \textit{et al.}~\cite{yin2019towards} & - & - & - & 0.1 & 83.8 \\
Xie \textit{et al.}~\cite{xie2018comparator} & - & - & 88.5 & 94.7 & 98.3 \\
Zhao \textit{et al.}~\cite{zhao2019look} & - & 82.6 & 89.5 & 93.5 & 96.2 \\
multicolumn~\cite{xie2018multicolumn} & - & 77.1 & 86.2 & 92.7 & 96.8 \\
VGGFace2~\cite{cao2018vggface2} & - & 74.7 & 84.0 & 91.0 & 96.0 \\
PFE~\cite{shi2019PFE} & - & 89.64 & 93.25 & 95.49 & 97.17 \\
ArcFace~\cite{deng2019arcface} & 86.25 & 93.15 & 95.65 & 97.20 & 98.18 \\ \hline
\textit{Average} & 86.69 & 92.72 & 94.89 & 96.62 & 97.90 \\
DDL & \textbf{92.39} &\textbf{ 94.89} & \textbf{96.41} & \textbf{97.47} & \textbf{98.33}\\ \hline
\end{tabular}%
}
\label{tab:ijbc}
\end{table}
To qualitatively evaluate the discriminability pattern learned by our DDL, we visualize the discriminability score distribution for two template images in IJB-C datasets. As shown in Figure \ref{fig:vis_ijbc}, DDL can effectively identify image discriminability. Images with large poses, visual blur, occlusion, and incomplete content are regarded to be low discriminative. The efficient discriminability judgment ability for our DDL leads to an extraordinary performance on set-to-set face recognition problems.

\subsubsection{Ablation Study}

~\newline

\textbf{The architecture of DDNet and base model.} In the above experiments, we have adopted the channel reduced version of ResNet-18 as the backbone for DDNet.  When inference, all test samples will be sent to DDNet firstly to predict discriminability. Therefore, the test computational cost is very sensitive to the architecture of DDNet. We design it as light-weight as possible. We also conduct experiments with wider and deeper DDNet and test on IJB-C. As shown in Table~\ref{tab:abalation_study_ijbc}, the wider and deeper networks have not brought significant performance gains.

As for the base model, we also experiment DDL with MobileFaceNet~\cite{chen2018mobilefacenets}, a popular backbone for mobile devices. From Table~\ref{tab:abalation_study_ijbc}, we can see that by combining will DDL, a consistent performance gain can be achieved on set-to-set face recognition task for MobileFaceNet.

\begin{table}[t]
\centering

\caption{ Ablation study with different DDNet architecture,  base model architecture and training datasets. Results are reported on IJB-C benchmark. 'CD' means channel reduced }
\resizebox{1\textwidth}{!}{%
\begin{tabular}{ccc|ccccc}
\hline
\multicolumn{3}{c|}{Method} & \multicolumn{5}{c}{IJB-C (TAR@FAR)} \\ \hline
DDNet & Base Model & Train Datasets & 1e-6 & 1e-5 & 1e-4 & 1e-3 & 1e-2 \\ \hline
ResNet-18-CD & ResNet-101 & MS-Celeb-1M & \textbf{91.14} & \textbf{95.75} & \textbf{96.94} & 97.72 & \textbf{98.36} \\
ResNet-34-CD & ResNet-101 & MS-Celeb-1M & 91.13 & 95.74 & 96.90 & 97.72 & 98.33 \\
ResNet-18 & ResNet-101 & MS-Celeb-1M & 90.93 & 95.74 & 96.92 & \textbf{97.73} & 98.35 \\ \hline
ResNet-18-CD & MobileFaceNet & MS-Celeb-1M & \textbf{87.32} & \textbf{91.45} & \textbf{94.30} & \textbf{96.24} & \textbf{97.82} \\
- & MobileFaceNet & MS-Celeb-1M & 79.88 & 88.21 & 92.08 & 95.22 & 97.24 \\ \hline
ResNet-18-CD & ResNet-101 & IMDB-Face & \textbf{88.35} & \textbf{92.26} & \textbf{95.09} & \textbf{96.71} & \textbf{98.05} \\
- & ResNet-101 & IMDB-Face & 73.12 & 86.44 & 92.44 & 94.70 & 97.40 \\ \hline
\end{tabular}%
}
\label{tab:abalation_study_ijbc}
\end{table}

\textbf{Train on other datasets.} In the aforementioned experiments, we use the MS-Celeb-1M dataset for the base model and DDNet training. To demonstrate the good generalization of our method, we also train the base model and DDNet with IMDB-Face~\cite{wang2018devil} dataset. IMDb-Face is a new large-scale noise-controlled dataset for face recognition. The dataset contains about 1.7 million faces, 59k identities, which is manually cleaned from 2.0 million raw images.  The results on IJB-C are shown in Table~\ref{tab:abalation_study_ijbc}, DDL improves set-to-set face recognition by a huge margin compared with simple average pooling, up to 15.23\% at FPR=1e-6. The model trained on IMDB-Face tends to be weaker than MS-Celeb-1M and more easily confused by hard negative pairs, thus DDL achieves a  more significant improvement.

\textbf{The influence of re-scale.} In Eq (9), we re-scale the discriminability scores of  element set between 0 and 1 to ensure the same range for element sets with different lengths. In this part, we compare the re-scale strategy and origin scale on  IQIYI-VID-FACE benchmark. As  shown in Table~\ref{tab:iqiyi},  re-scale can boost performance for 1.67\%. For video face recognition, which contains various frames from dozens of to thousands of, it is necessary to re-scale the predicted discriminability scores at the test stage.

\textbf{Combined with more loss functions.} There are many successful loss function these years for face recognition task, such as ArcFace~\cite{deng2019arcface}, CosFace~\cite{wang2018cosface}  and SphereFace~\cite{liu2017sphereface}. We combine DDL with more loss functions and test on YouTube Face benchmark. As shown in Table~\ref{tab:abaltion_ytf}, all loss functions achieve constant performance gain with DDL. DDL is not sensitive to the base model training loss function and can easily cooperate with any existing base.

\begin{table}[t]
\caption{Ablation study with loss function. Results are reported on YouTube Face benchmark}
\centering
\begin{tabular}{ccc}
\hline
\multicolumn{2}{c}{Method}                & \multirow{2}{*}{Accuracy(\%)}   \\ \cline{1-2}
DDL                       & loss function &                                 \\ \hline
\checkmark & ArcFace       & \textbf{98.18} \\
$\times$                  & ArcFace       & 97.97                           \\ \hline
\checkmark & CosFace       & \textbf{97.91} \\
$\times$                  & CosFace       & 97.68                           \\
\checkmark & SphereFace    & \textbf{97.12} \\
$\times$                  & SphereFace    & 96.83                           \\ \hline
\end{tabular}

\label{tab:abaltion_ytf}
\end{table}

\subsection{Video-Based Person Re-Identification}

In this section, we will evaluate our DDL with the video-based person re-identification task on Mars~\cite{zheng2016mars}. It the largest video-based person re-identification dataset. The train and test set are followed official split.

To train the base model, triplet loss function and softmax cross-entropy loss function are used. The similarity metric is  L2 distance.  Standard ResNet-50 pre-trained on ImageNet is used and video frames are resized to 224$\times$112. We will report mean average precision score (mAP) and cumulative matching curve (CMC) at rank-1, rank-5 and rank-20. Note that re-rank is not applied in the comparison.

The results are shown in Table~\ref{tab:this_mars}, DDL boosts the performance consistently. Compared with average pooling, DDL achieves performance  gain for 3.6\% mAP. For more complicated aggregation strategies like RNN and the state-of-the-art attention mechanism, DDL also improves performance. The good performance of video-based person re-identification further demonstrates the efficiency of our DDL in group representation learning.

\begin{table}[t]

\caption{Results for video-based person re-identification on Mars} 

\setlength{\tabcolsep}{12pt}
\centering
\begin{tabular}{ccccc}
\hline
                      & mAP  & CMC-1 & CMC-5 & CMC-20 \\ \hline
Zheng \textit{et al.}\cite{zheng2016mars}          & 45.6 & 65.0  & 81.1  & 88.9   \\
Li \textit{et al.} \cite{li2017learning}            & 56.1 & 71.8  & 86.6  & 93.1   \\
QAN \cite{liu2017quality}            & 51.7 & 73.7  & 84.9  & 91.6   \\
Hermans\textit{ et al.}\cite{hermans2017defense}        & 67.7 & 79.8  & 91.4  & -      \\
 3D conv ~\cite{gao2018revisiting}    & 70.5 & 78.5  & 90.9  & 95.9   \\
 Atttention~\cite{gao2018revisiting} & 76.7 & 83.3  & 93.8  & 97.4   \\
RNN ~\cite{gao2018revisiting}        & 73.9 & 81.6  & 92.8  & 96.3   \\ \hline
\textit{average}            & 74.1 & 81.3  & 92.6  & 96.7   \\
DDL                  & \textbf{77.7} & \textbf{84.0}  & \textbf{94.8}  & \textbf{97.4}   \\ \hline
\end{tabular}

\label{tab:this_mars}
\end{table}

\subsection{Action Recognition}

In this section, we will evaluate our DDL on two most popular action recognition datasets ActivityNet-1.2~\cite{caba2015activitynet} and Kinetics-700~\cite{kay2017kinetics}. The ActivityNet-1.2 contains 4,819 training videos and 2,383 validation videos for 100 action class. It is an untrimmed video dataset, namely more temporal variance and noises there are. The Kinetics-700 is a well-trimmed action recognition datasets, which contains over 650k videos from 700 classes.

All video frames are extracted by FFmpeg with 30fps then resized and center crop to 112$\times$112. We select three clip-based action recognition baseline method, the 3D-ResNet-50~\cite{hara2018can}, SlowFast-50~\cite{feichtenhofer2018slowfast} and R(2+1)D-50~\cite{tran2018closer}. The training config for those base models follows SlowFast~\cite{feichtenhofer2018slowfast}. In the original approach, all three methods rely on dense sampling during testing. To be more specific, they oversampling both spatially and temporally to capture target activation. 

 The DDNet architecture for action recognition is the same with image task, but replace all 2D-Conv to 3D-Conv.  A video will firstly be divided into many clips, and each clip's discriminability will be generated by DDNet, only top-K clips will be extracted feature and aggregated. The K we select for ActivityNet-1.2 and Kinetics is 9 and 5, respectively. We select random and uniform sampling K clips for comparison with sampling by DDL. A dense sampling experiment is also conducted.
 
From Table~\ref{tab:activitynet12}, DDL improves recognition performance for all baseline models on ActivityNet-1.2. For the state-of-the-art clips-based model SlowFast, combining it with DDL can achieve around 4\% accuracy gain compared with random or uniform sampling on ActivityNet-1.2. What's more, DDL can even outperform dense sampling by 2.49\%, while the dense sampling strategy sample above 5x more clips (estimated by the average duration 120s for ActivitNet-1.2).

For Kinetics-700, the results are in Table~\ref{tab:kinetics700}. DDL outperforms random sampling by 1.84\% and uniform sampling by 2.18\%. For dense sampling, DDL can achieve 0.46\% gain with 6x speed up. Since the Kinetics-700 is trimmed by human and video quality is under control, combining with DDL can also significantly boost recognition performance and save computational consumption.

\begin{table}[t]

\caption{ Video action recognition results(\%) on ActivityNet-1.2 dataset. Accuracy is reported by top-1 on the validation set}
\centering
\begin{tabular}{ccccc}
\hline
Model & DDL & Random & Uniform & Dense \\ \hline
clip number & \multicolumn{3}{c}{9} & 60 \\ \hline
3D-RS-50 & \textbf{86.38} & 82.83 & 83.14 & 83.92 \\
R(2+1)D-RS-50 & \textbf{89.08} & 84.51 & 84.89 & 85.46 \\
SlowFast-RS-50 & \textbf{90.21} & 85.92 & 86.14 & 87.72 \\ \hline
\end{tabular}
\label{tab:activitynet12}
\end{table}

\begin{table}[t]

\caption{Video action recognition results(\%) on Kinetics-700 dataset.  Accuracy is reported on the validation set and is the average of top1 and top 5 accuracy}
\centering
\begin{tabular}{ccccc}
\hline
Model & DDL & Random & Uniform & Dense \\ \hline
clip number & \multicolumn{3}{c}{5} & 30 \\ \hline
3D-RS-50 & \textbf{71.01} & 68.26 & 67.43 & 68.83 \\
R(2+1)D-RS-50 & \textbf{72.51} & 69.24 & 68.79 & 70.94 \\
SlowFast-RS-50 & \textbf{74.23} & 72.39 & 72.05 & 73.77 \\ \hline
\end{tabular}
\label{tab:kinetics700}
\end{table}

\section{Conclusion}

In this paper, we have proposed a novel post-processing module called Discriminability Distillation Learning (DDL) for all group-based recognition tasks. We explicitly define the discriminability with observations on feature embedding, then apply a light-weight network for discriminability distillation and feature aggregation. We identify the advantage of our proposed methods in the following aspects: (1) The entire discriminability distillation is performed without modifying the pre-trained based network, which is highly flexible comparing with existing quality-aware or attention methods. (2) Our distillation network is extremely light-weighted which saves great computational cost. (3) With our DDL and feature aggregation, we achieve state-of-the-art results on multiple group-based recognition tasks including set-to-set face recognition, video-based person re-identification, and action recognition.

\bibliographystyle{splncs04}
\bibliography{egbib}
\end{document}